\documentclass[11pt]{article}

\usepackage[preprint]{acl}

\usepackage{times}
\usepackage{latexsym}

\usepackage[T1]{fontenc}

\usepackage[utf8]{inputenc}
\usepackage{microtype}
\usepackage{inconsolata}
\usepackage{graphicx}
\usepackage{makecell}
\usepackage{amsmath}
\usepackage{booktabs}
\usepackage{multicol}
\usepackage{multirow}
\usepackage{amssymb}
\usepackage{pifont}
\newcommand{\xmark}{\ding{55}} %
\usepackage{tabularx}
\usepackage{array}

\usepackage{listings}
\usepackage{xcolor}
 
\title{Precision–Recall Trade-off on the Inference Level for LLMs in Grammatical Error Correction}

\author{Ryszard Staruch \\
  Adam Mickiewicz University \\
  \texttt{ryszard.staruch@amu.edu.pl} \\\And
  Filip Graliński \\
  Adam Mickiewicz University \\
  \texttt{filip.gralinski@amu.edu.pl} \\}

\begin{document}
\maketitle
\begin{abstract}

Large language models (LLMs) have recently been adopted for Grammatical Error Correction (GEC), achieving strong performance via both full fine-tuning and prompt-based approaches. As probabilistic sequence models, LLMs expose inference-time parameters that directly modify token generation behavior; however, their potential for controlling GEC-specific trade-offs has not been systematically studied. In this work, we investigate decoding strategies for grammatical error correction. We show that the repetition penalty parameter enables controllable adjustment of the precision–recall trade-off at inference time, without modifying the model architecture, training procedure, or prompt. Through experiments on standard GEC benchmarks, we demonstrate that repetition-penalty-based decoding enables inference-time control of the precision–recall balance, conceptually related to methods previously proposed for sequence-to-sequence GEC models, such as Align-and-Predict Decoding (APD), while remaining simple and broadly applicable to existing LLMs that expose inference-time decoding penalties.

\end{abstract}

\section{Introduction}

Grammatical Error Correction (GEC) is a field in which systems are expected to identify and correct errors in texts. The task is fairly subjective -- it depends on the context in which the system operates, and its goals can differ. For example, a system assisting language learners should provide fewer but more precise corrections to avoid harming or misteaching the learner~\cite{ng-etal-2014-conll}. In contrast, a GEC system for native speakers should find as many errors as possible, since most of them stem from omissions~\cite{wang-etal-2022-cctc}. The system can thus be optimized for high precision or high recall, balancing the two objectives in a precision–recall trade-off. Table~\ref{tab:gec_example} illustrates the differences between precision-oriented and recall-oriented GEC systems.

\begin{table}[t]
\normalsize
\centering
\begin{tabularx}{\columnwidth}{@{}>{\raggedright\arraybackslash}X@{}}
\hline
\textbf{Source text} \\
for example, in math club, teachers usually teach us how to solve a math problem using vary ways. \\
\hline
\textbf{Precision-oriented GEC} \\
\textbf{For} example, in math club, teachers usually teach us how to solve a math problem using \textbf{various} ways. \\
\hline
\textbf{Recall-oriented GEC} \\
\textbf{For} example, in \textbf{the} math club, teachers usually teach us how to solve a math problem using \textbf{various methods}. \\
\hline
\end{tabularx}
\caption{Example illustrating the difference between precision-oriented and recall-oriented GEC. Precision-oriented GEC is more conservative and tends to make only minimal, high-confidence edits, whereas recall-oriented GEC is less strict and proposes a broader set of changes, including optional stylistic improvements.}
\label{tab:gec_example}
\end{table}

In the pre-LLM era, the most effective approaches were sequence-to-edit (seq2edit) and sequence-to-sequence (seq2seq) models. The former tag the incorrect text and apply changes according to the tags~\cite{omelianchuk-etal-2020-gector}, while the latter directly return the corrected version of the text~\cite{junczys-dowmunt-etal-2018-approaching}.
 
LLMs showed decent performance on GEC, but tend to produce overcorrected outputs under standard greedy decoding \cite{omelianchuk-etal-2024-pillars, davis-etal-2024-prompting}. Their behavior can be controlled by modifying the decoding strategy via inference parameters such as top-$k$, top-$p$, temperature, or penalty mechanisms. Such parameters can improve performance, although the decoding strategy must be carefully chosen for the task \cite{wiher-etal-2022-decoding}. Unfortunately, this has not been explored for GEC.
 
To address this gap, we analyze the available inference parameters in the context of GEC to identify those that can benefit LLMs adapted to the task. Our experiments show that the repetition penalty can successfully control the precision–recall trade-off. We analyze how this parameter behaves for both a general instruction-tuned LLM and an LLM fine-tuned for the task, and whether using a prompt with the fine-tuned LLM affects the decoding strategy under the repetition penalty and the overall performance.
 
Finally, we provide practical guidance on using the repetition penalty parameter across different contexts in which GEC systems can be applied.

\section{Precision–recall Trade-off in GEC}

The precision–recall trade-off characterizes how a correction model behaves. Models tuned toward higher precision act conservatively, producing fewer corrections, while recall-oriented models propose more changes and may introduce unnecessary edits.
 
For seq2edit models, the balance can be adjusted at inference time through a bias toward the \texttt{KEEP} tag and a minimum error probability threshold. For seq2seq models, \citet{sun-wang-2022-adjusting} propose Align-and-Predict Decoding (APD), which intervenes in decoding based on alignment with the source text, enabling a single model to produce more conservative or more aggressive corrections without retraining.
 
A separate line of work achieves similar control through post-processing. \citet{sorokin-2022-improved} reranks the elementary edits of a first-stage model with a dedicated classifier and applies only those above a threshold; raising it discards unreliable edits and substantially increases precision, while recall stays bounded by the first-stage model's coverage. \citet{qorib-ng-2023-system} enlarge the candidate pool by scoring complete hypotheses built from the edits of several base systems, at the cost of running multiple systems. Similarly, \citet{park-etal-2025-leveraging} propose Post-Correction via Overcorrection (PoCO), which uses smaller models to filter edits from over-correcting LLMs. Such methods offer largely one-sided control and add both a second model and extra latency.
 
Other work adapts LLMs to GEC directly: \citet{zeng-etal-2024-evaluating} analyze prompting strategies, while \citet{staruch-etal-2025-adapting} and \citet{liang-etal-2025-edit} investigate training-time adaptation under precision-oriented evaluation. Prompting, however, is unstable and insufficient for mitigating overcorrection, and training-based approaches require substantial compute and access to model weights.
 
These limitations motivate our work on inference-time control of the precision–recall trade-off in LLM-based GEC. Unlike APD, which requires a custom decoding implementation, and post-correction pipelines, which add processing stages, our approach relies only on inference parameters exposed by standard LLM decoding interfaces, making it broadly compatible with existing deployment frameworks.

\begin{table*}[t]
\centering
\small
\begin{tabular}{lcc}
\hline
\textbf{Property} & \textbf{APD} & \textbf{Repetition Penalty} \\
\hline
Inference-time method & \checkmark & \checkmark \\
Score modification & Multiplicative & Multiplicative \\
Score space & Log-probabilities & Logits \\
Source awareness & Explicit alignment & Prompt token presence \\
Granularity & Position-aware & Token-identity-based \\
Targets only the aligned source token & \checkmark & \xmark \\
LLM API Compatibility & \xmark & \checkmark \\

\hline
\end{tabular}
\caption{Comparison of APD method and repetition-penalty-based decoding strategy.}
\label{tab:apd_vs_repetition}
\end{table*}

\section{Decoding Strategies for LLM-based Grammatical Error Correction}

Different tasks require task-specific decoding strategies when using large language models. GEC is a constrained generation task in which the model should introduce necessary corrections while preserving the structure and meaning of the source text. This favors conservative generation behavior and motivates careful tuning of inference-time decoding parameters.

\subsection{LLM Inference Parameters}

The available inference parameters depend on the deployment framework. In this work, we focus on vLLM~\cite{10.1145/3600006.3613165}, a widely used open-source library for efficient LLM inference.

Temperature controls the randomness of token generation. Since GEC expects a single high-confidence correction, we set it to zero. Beam search could further improve sequence likelihood, but it significantly increases inference time and is therefore not used. For the same reason, we do not consider stochastic sampling strategies such as top-$p$, top-$k$, or min-$p$: they are intended to promote output diversity, which is inherently unsuitable for error correction, as selecting tokens beyond the most probable ones lacks justification without additional information.

Common token-level penalties include the frequency, presence, and repetition penalties. The frequency penalty grows with the number of times a token has already appeared in the generated output, while the presence penalty penalizes tokens based solely on whether they have appeared at least once. Both apply only to tokens generated during inference and ignore prompt tokens, so they cannot directly discourage copying from the source text in GEC.

In contrast, the repetition penalty also takes into account tokens present in the prompt, allowing it to be applied to all tokens, including those in the original uncorrected text.

Let $z_t \in \mathbb{R}$ denote the logit of token $t$ and let $r > 0$ be the repetition penalty.
We define an indicator function
\[
\mathbb{I}(t) =
\begin{cases}
1 & \text{if } t \in \text{prompt} \cup \text{current output}, \\
0 & \text{otherwise}.
\end{cases}
\]

The repetition-penalized logit $z'_t$ is given by
\[
z'_t =
\begin{cases}
\dfrac{z_t}{r} & \text{if } z_t > 0 \text{ and } \mathbb{I}(t)=1, \\
z_t \cdot r & \text{if } z_t \le 0 \text{ and } \mathbb{I}(t)=1, \\
z_t & \text{if } \mathbb{I}(t)=0.
\end{cases}
\]

Since the penalty modifies logits multiplicatively, values of $r$ greater than 1 discourage reusing tokens already seen in the prompt or output, whereas values below 1 encourage such repetition (effectively becoming a repetition reward).

In GEC, the penalty adjusts the probabilities of all source-text tokens and thus influences generation. Since introducing changes usually involves adding new tokens, the penalty can steer the model toward a more conservative output or one with more extensive changes.

Its advantage is ease of use: the penalty is set through a single argument in standard inference interfaces -- example code is provided in Appendix~\ref{sec:appendix}.

\subsection{Repetition Penalty Comparison to the APD Method}

Applying a repetition penalty during LLM inference is conceptually related to the APD method; Table~\ref{tab:apd_vs_repetition} compares the two. Both use multiplicative transformations to adjust token scores, but in different score spaces: the repetition penalty is applied directly to logits, whereas APD modifies log probabilities.

A key distinction is that APD operates on the exact token selected at each decoding step, while the repetition penalty considers all tokens present in the prompt and the current output at every step. The repetition penalty thus offers a simpler alternative to APD, at the cost of adjusting all previously seen tokens rather than only the aligned source token. When the source text contains mostly unique tokens and corrections introduce mainly novel tokens, we hypothesize the practical difference is small; a direct empirical comparison, however, requires a custom APD implementation for autoregressive LLMs and is left for future work.

\section{Experiments}

\begin{figure*}[t]
    \centering
    \includegraphics[width=\textwidth]{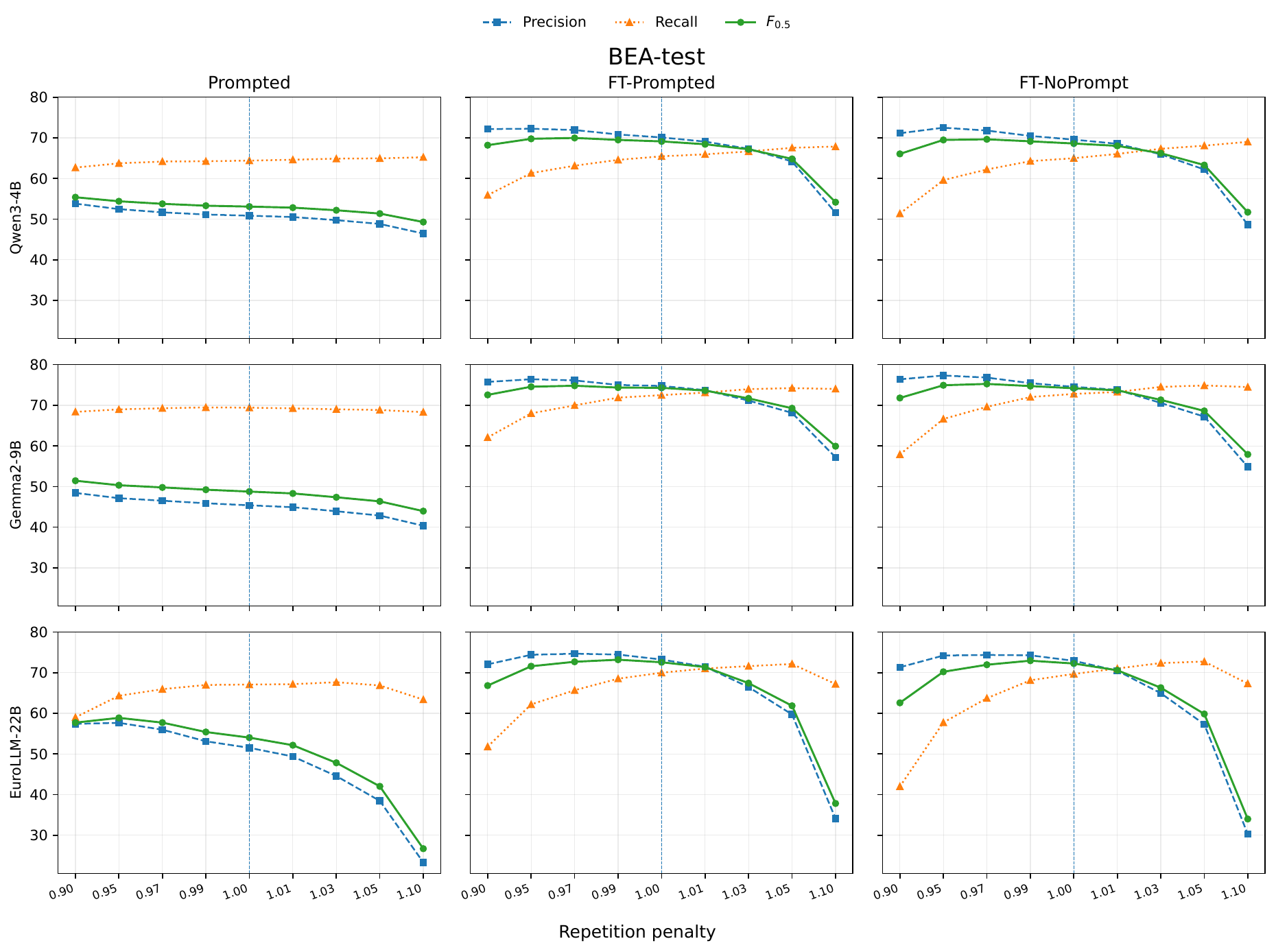}
    \caption{
    Effect of repetition penalty on English (BEA-test) GEC benchmark.
    A repetition penalty value of 1 indicates no penalty.
    }
    \label{fig:rep_penalty_eng}
\end{figure*}

To investigate the impact of the repetition penalty parameter on LLM inference for the GEC task, we conducted experiments on three benchmarks: BEA-test (W\&I+LOCNESS)~\cite{bryant-etal-2019-bea} for English, Falko-MERLIN~\cite{boyd2018wnut} for German, and UA-GEC~\cite{syvokon-etal-2023-ua} for Ukrainian. For more comprehensive results, we additionally report performance on two common English benchmarks, CoNLL-2014~\cite{ng-etal-2014-conll} and JFLEG~\cite{napoles-etal-2017-jfleg}, in Tables~\ref{tab:rep_penalty_conll_bea_jfleg_qwen}--\ref{tab:rep_penalty_conll_bea_jfleg_eurollm} in the Appendix. Dataset statistics are given in Table~\ref{tab:dataset_stats}.

To assess its applicability to different LLMs, we use the following models:
\begin{itemize}
    \item Qwen3-4B,\footnote{\url{https://huggingface.co/Qwen/Qwen3-4B-Instruct-2507}}
    \item Gemma2-9B,\footnote{\url{https://huggingface.co/google/gemma-2-9b-it}}
    \item EuroLLM-22B.\footnote{\url{https://huggingface.co/utter-project/EuroLLM-22B-Instruct-2512}}
\end{itemize}

The model selection criteria were based on general public interest, model size, multilingual capabilities, country of origin and license openness.

To analyze different usage scenarios, we evaluate the models under three configurations:
\begin{itemize}
\item \textbf{*-Prompted}: a model without task-specific fine-tuning, evaluated using a detailed prompt that explicitly describes the GEC task.
\item \textbf{*-FT-Prompted}: a version of the model fine-tuned for GEC and trained with the detailed task prompt format; evaluation is performed using the same detailed prompt.
\item \textbf{*-FT-NoPrompt}: a separate version of the model fine-tuned for GEC without explicit task prompts; during evaluation, only the input text to be corrected is provided in the user part of the chat template.
\end{itemize}

Comparing *-Prompted with *-FT-Prompted allows us to assess how changes in the repetition penalty affect a general-purpose LLM versus a model fine-tuned for GEC. In turn, the comparison between *-FT-Prompted and *-FT-NoPrompt examines whether tokens originating from the task description prompt influence the behavior of the repetition penalty in the GEC setting. It also allows us to examine whether using a prompt for the GEC task is necessary when the model has been fine-tuned.

\begin{figure*}[p]
    \centering
    \includegraphics[width=\textwidth]{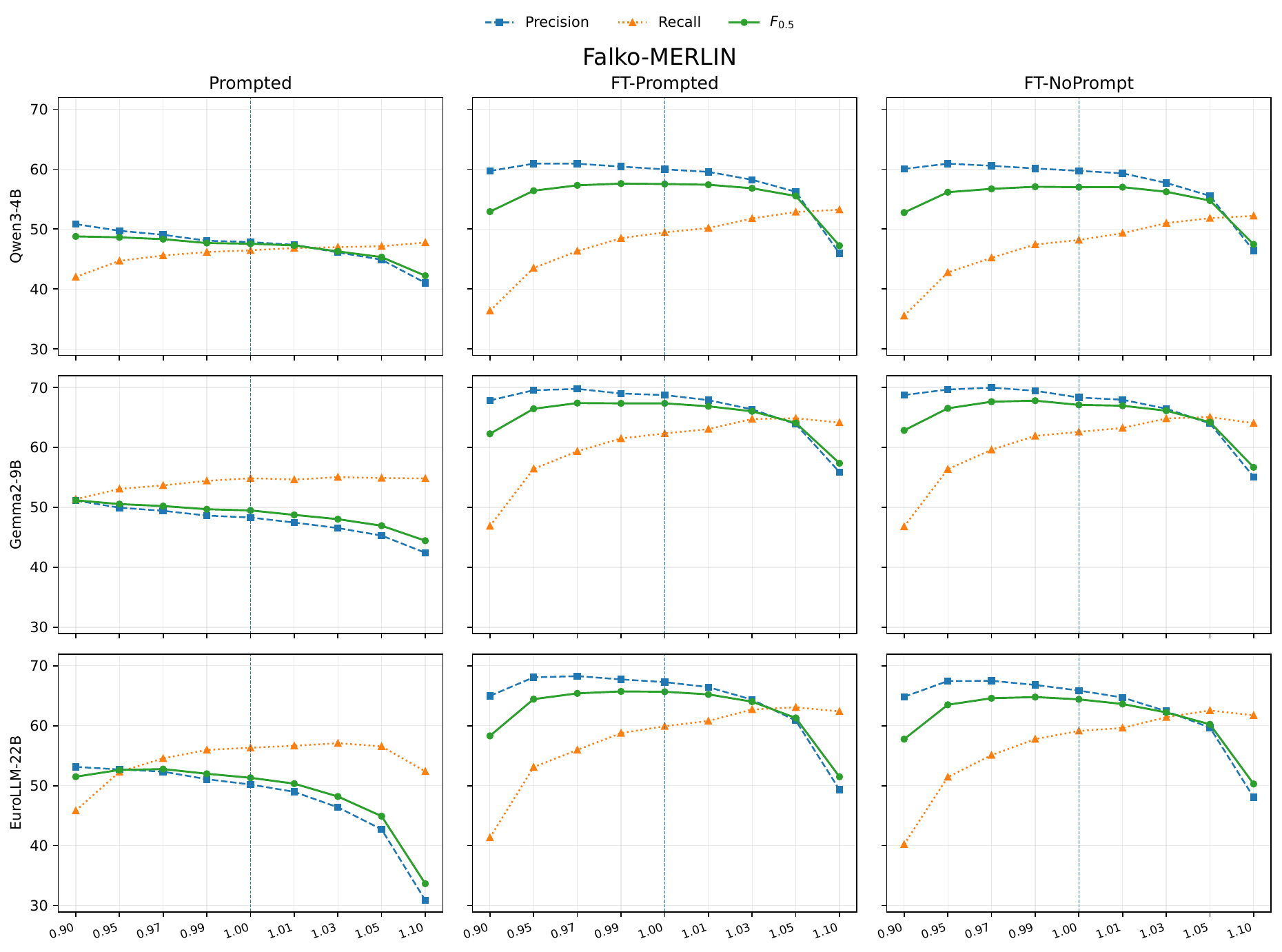}
    \vspace{1mm}
    \includegraphics[width=\textwidth]{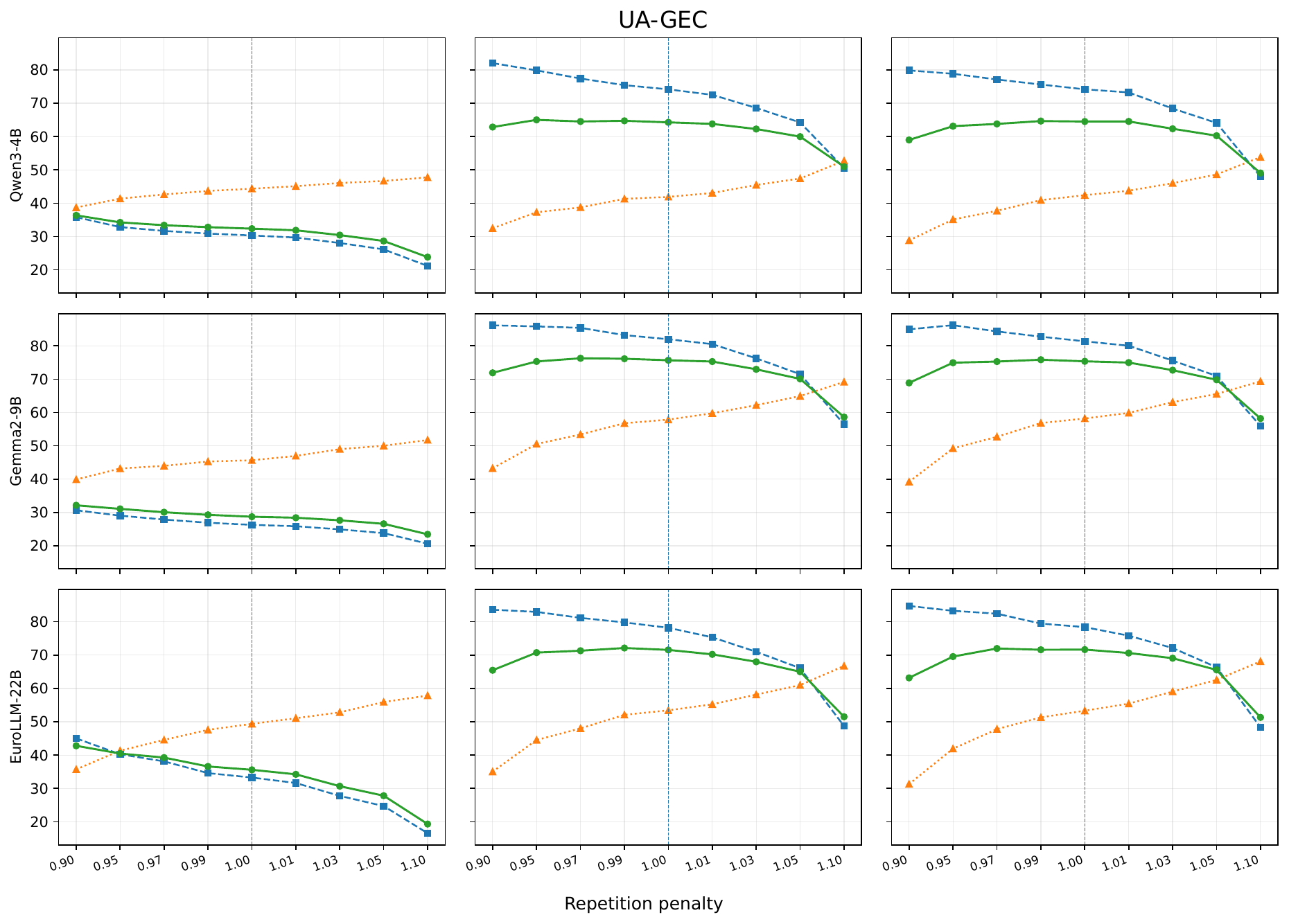}
    \caption{
    Effect of repetition penalty on German (Falko-MERLIN) and Ukrainian (UA-GEC) GEC benchmarks.
    A repetition penalty value of 1 indicates no penalty.
    }
    \label{fig:rep_penalty_multilingual}
\end{figure*}

\begin{table*}[t]
\centering
\small
\begin{tabular}{lccccccccc}
\toprule
\multirow{2}{*}{\textbf{Model}} & \multicolumn{3}{c}{\textbf{BEA-test}} & \multicolumn{3}{c}{\textbf{Falko-MERLIN}} & \multicolumn{3}{c}{\textbf{UA-GEC}} \\
\cmidrule(lr){2-4} \cmidrule(lr){5-7} \cmidrule(lr){8-10}
& RP & $\Delta$F\textsubscript{0.5} & $\Delta$P & RP & $\Delta$F\textsubscript{0.5} & $\Delta$P & RP & $\Delta$F\textsubscript{0.5} & $\Delta$P \\
\midrule
Qwen3-4B-Prompted & 0.90$^\dagger$ & +2.30 & +2.97 & 0.90$^\dagger$ & +1.22 & +2.99 & 0.90$^\dagger$ & +3.99 & +5.49 \\
Qwen3-4B-FT-Prompted & 0.97 & +1.04 & +2.23 & 0.99 & +0.07 & +0.47 & 0.95 & +0.74 & +5.70 \\
Qwen3-4B-FT-NoPrompt & 0.97 & +0.85 & +1.82 & 0.99 & +0.07 & +0.40 & 0.99 & +0.14 & +1.44 \\
\midrule
Gemma2-9B-Prompted & 0.90$^\dagger$ & +2.67 & +3.05 & 0.90$^\dagger$ & +1.70 & +2.83 & 0.90$^\dagger$ & +3.43 & +4.38 \\
Gemma2-9B-FT-Prompted & 0.97 & +0.51 & +1.36 & 0.97 & +0.05 & +1.03 & 0.97 & +0.59 & +3.42 \\
Gemma2-9B-FT-NoPrompt & 0.97 & +1.07 & +2.27 & 0.99 & +0.70 & +1.12 & 0.99 & +0.47 & +1.40 \\
\midrule
EuroLLM-22B-Prompted & 0.95 & +4.86 & +6.15 & 0.97 & +1.46 & +2.15 & 0.90$^\dagger$ & +7.20 & +11.75 \\
EuroLLM-22B-FT-Prompted & 0.99 & +0.70 & +1.35 & 0.99 & +0.05 & +0.45 & 0.99 & +0.56 & +1.62 \\
EuroLLM-22B-FT-NoPrompt & 0.99 & +0.63 & +1.23 & 0.99 & +0.38 & +0.94 & 0.97 & +0.33 & +4.02 \\
\bottomrule
\end{tabular}
\caption{Repetition penalty value maximizing F\textsubscript{0.5} (RP) and the corresponding gains in F\textsubscript{0.5} and precision over no penalty (RP $=1.0$) for each model configuration and benchmark. $\Delta$P is the precision change at the same RP that maximizes F\textsubscript{0.5}, not at the precision-maximizing RP. $\dagger$: optimum lies at the boundary of the tested range ($0.90$); lower values were not evaluated and may perform better.}
\label{tab:f05-argmax}
\end{table*}

All experiments are performed using greedy decoding with a repetition penalty applied. The temperature is set to 0, and beam search is not used, as this configuration is the most practical for real-time LLM deployment. Details of the prompt design and the fine-tuning procedure are provided in the Appendix.

The evaluation for BEA-test is conducted using ERRANT~\cite{bryant-etal-2017-automatic} on the CodaBench platform, as the benchmark labels are not publicly available. The Falko-MERLIN and UA-GEC benchmarks were evaluated using ERRANT within the gec-metrics tool \cite{goto-etal-2025-gec}. CoNLL-2014 and JFLEG were evaluated using their official scorers.

We note, however, that our experiments are not intended to tune the model toward precision- or recall-oriented evaluation as measured by the F\textsubscript{$\beta$}-score for a particular $\beta$. Different scenarios and application contexts may require different $\beta$ values, so we do not optimize specifically for F\textsubscript{${0.5}$}, even though it is reported as the primary metric in the evaluated benchmarks. Instead, our goal is to demonstrate that the precision–recall trade-off can be robustly controlled via the repetition penalty across different models and benchmarks. In particular, we show that increasing the penalty above 1.0 consistently steers the model toward producing more changes, whereas lowering it below 1.0 results in stricter behavior.

\begin{table*}[t]
\centering
\begin{tabular}{lrr|lrr}
\toprule
\multicolumn{3}{c}{Top $\Delta$Precision (repetition penalty=0.95)} & \multicolumn{3}{c}{Top $\Delta$Recall (repetition penalty=1.05)} \\
Error & $\Delta P$ & $\Delta R$ & Error & $\Delta P$ & $\Delta R$ \\
\midrule
ADJ & 15.34 & -6.30 & VERB & -10.99 & 8.27 \\
PART & 14.41 & -10.75 & NOUN:NUM & -9.40 & 6.90 \\
CONJ & 13.33 & -18.10 & PRON & -8.94 & 6.90 \\
NOUN & 6.32 & -9.80 & NOUN:POSS & -1.33 & 5.81 \\
VERB:TENSE & 5.33 & -11.72 & ADV & -11.95 & 5.75 \\
NOUN:NUM & 4.50 & -8.41 & NOUN & -5.49 & 5.04 \\
VERB:SVA & 4.34 & -3.53 & ADJ & -9.02 & 4.94 \\
ORTH & 3.69 & -7.20 & VERB:TENSE & -9.22 & 3.97 \\
PREP & 3.58 & -10.96 & ORTH & -3.20 & 3.69 \\
VERB:FORM & 3.43 & -3.87 & DET & -2.63 & 1.99 \\
\bottomrule
\end{tabular}
\caption{Precision–recall differences for the Gemma2-9B-FT-NoPrompt model across error types on the BEA-test set. The left part reports precision and recall changes at repetition penalty 0.95 compared with standard decoding, while the right part reports changes at repetition penalty 1.05. Only the ten most affected error types with at least 20 gold instances are shown.}
\label{tab:errortypes}
\end{table*}

\section{Results}

\subsection{Quantitative Results}
Figures~\ref{fig:rep_penalty_eng} and~\ref{fig:rep_penalty_multilingual}
present the precision, recall, and F\textsubscript{0.5} curves for
different repetition penalty values -- exact values are provided in
Tables~\ref{tab:rep_penalty_conll_bea_jfleg_qwen}--\ref{tab:rep_penalty_de_ua_eurollm} in the Appendix. A consistent trend holds across all models and benchmarks:
penalty values below 1 increase precision while reducing recall, whereas
values above 1 lower precision and raise recall, in line with the
expected behavior of the penalty. Extreme values, however, can degrade both metrics simultaneously; within 0.97–1.03 this was rare, occurring only for some non–fine-tuned configurations (e.g., Gemma2-9B-Prompted on BEA-test). Metric changes are also smoother for the
non--fine-tuned models than for the fine-tuned variants, possibly
because fine-tuning alters the magnitude or distribution of logits and
thereby amplifies the effect of logit-level penalties.

The trends of the *-FT-Prompted and *-FT-NoPrompt models are highly
similar, suggesting that task-specific instructions in the prompt have
limited influence on the penalty's behavior and that fine-tuned models
can operate effectively without explicit instructional prompts, as the
task behavior appears to be largely encoded in the model parameters.

\subsection{Precision-Oriented Analysis}
\label{sec:precision-analysis}
While we do not optimize for any particular
F\textsubscript{$\beta$} score, the benchmarks report
F\textsubscript{0.5} as their primary metric, so we examine how the
penalty affects it. Table~\ref{tab:f05-argmax} reports, for each model
configuration and benchmark, the penalty value that maximizes
F\textsubscript{0.5} and the corresponding gains over standard
decoding. In all 27 cases in Table~\ref{tab:f05-argmax} the optimum lies below 1.0, indicating that the penalty's direction of action is systematically aligned with precision across models, adaptation regimes, and languages.\footnote{On CoNLL-2014 (Appendix), the same pattern holds except for two fine-tuned configurations whose optimum lies exactly at 1.0, within 0.2 F\textsubscript{0.5} of the sub-1.0 values -- consistent with the flat region around the optimum discussed below.} The two model groups behave differently: fine-tuned
configurations peak in a narrow 0.95--0.99 range, most often at
0.97--0.99, with modest gains (up to +1.07 F\textsubscript{0.5}),
whereas non--fine-tuned configurations peak at lower values
(0.90$^\dagger$--0.97) with substantially larger gains (up to +7.20 for
EuroLLM-22B on UA-GEC), consistent with their stronger overcorrection;
in seven of nine such cases the optimum lies at the boundary of the
tested range, so even lower values may be beneficial.

Notably, the F\textsubscript{0.5} curves are flat in a wide region
around the optimum: the penalty substantially shifts the
precision--recall composition of the score, with precision gains of up
to +5.70 points for fine-tuned models and up to +11.75 points for
non--fine-tuned models at the F\textsubscript{0.5}-optimal value, while
the aggregate metric changes far less. Precision-oriented adjustment is
therefore available at negligible F\textsubscript{0.5} cost;
conversely, recall-oriented adjustment above 1.0 appears as mild
F\textsubscript{0.5} degradation only because of the metric's precision
weighting. The width of this flat region varies by model: EuroLLM-22B
exhibits a markedly narrower plateau and larger metric swings than the
other models in all setups, so the penalty value must be selected more
carefully in each individual case.

\begin{table*}
  \centering
  \begin{tabular}{p{15cm}}
    \hline
     \textbf{Source text:} Unforturntly, almost older people can not use internet, in spite of benefit of internet. \\
     \textbf{Standard decoding:} \textbf{Unfortunately}, \textbf{most} older people can not use the internet, in spite of the benefits of the internet.  \\
    \textbf{Repetition penalty=0.95:} \textbf{Unfortunately}, \textbf{almost all} older people can not use the internet, in spite of the benefits of the internet. \\
    \textbf{Repetition penalty=1.05:} \textbf{Unfortunately}, \textbf{most} older people can not use the internet, \textbf{despite} the benefits of the internet.  \\
    \hline
    \textbf{Source text:} Let's take time to compare the criminal life to the life of the hard working person. \\
     \textbf{Standard decoding:} Let's take time to compare the criminal life \textbf{with} the life of the \textbf{hard-working} person.  \\
    \textbf{Repetition penalty=0.9:} Let's take time to compare the criminal life to the life of the \textbf{hard-working} person. \\
    \textbf{Repetition penalty=1.1:} Let's take time to compare the criminal \textbf{lifestyle with that} of the \textbf{hardworking} person.  \\
    \hline
  \end{tabular}
  \caption{Examples of repetition penalty parameter impact on English GEC task during inference.}
  \label{tab:examples}
\end{table*}

\subsection{Impact on Error Types}
To examine which error types drive the trade-off, we compare the
Gemma2-9B-FT-NoPrompt model (one of the best-performing models) against
standard decoding on BEA-test: precision at a repetition penalty of
0.95 and recall at 1.05. We report only error types with at least 20
gold instances ($TP + FN \geq 20$), as categories with lower support
exhibit high variance.

Table~\ref{tab:errortypes} shows the most affected error types. As
expected, none gains in both precision and recall simultaneously. The
largest precision gains at 0.95 are observed for ADJ, PART, and CONJ
(each exceeding 10 percentage points), with notably smaller
improvements elsewhere; precision gains are not clearly correlated with
recall losses, as some types lose more recall than they gain in
precision. Recall gains at 1.05 are smaller overall, led by VERB,
NOUN:NUM, and PRON. Several types (ADJ, NOUN, VERB:TENSE, NOUN:NUM,
ORTH) appear in both lists, indicating particular sensitivity to the
penalty, but no single type is consistently and disproportionately
affected. We report this analysis for a single model; whether these
patterns generalize across models remains to be verified.

\subsection{Qualitative Analysis}
Table~\ref{tab:examples} shows examples of how the repetition penalty
affects model outputs for English sentences. The most obvious
correction, changing \textbf{Unforturntly} to \textbf{Unfortunately},
is produced consistently across all configurations. In the first
example, lowering the penalty causes the model to retain the word
\textbf{almost}, compensating by inserting \textbf{all} afterwards to
preserve grammaticality; only the highest penalty leads the model to
replace \textbf{in spite of} with \textbf{despite}.
In the second example, lowering the penalty leaves \textbf{to}
uncorrected (instead of \textbf{with}), yielding more conservative
edits, whereas the increased penalty produces a fluency-oriented
rewrite, replacing \textbf{criminal life} with \textbf{criminal
lifestyle with that of}. From a minimal-edit perspective such changes
may be considered unnecessary, illustrating how higher penalty values
encourage broader paraphrasing and thereby contribute to the observed
precision--recall trade-off. The use of \textbf{hardworking} instead of
\textbf{hard-working} is most likely due to the presence of this token
in the model's vocabulary.

\subsection{Practical Recommendations}
Based on the results analysis, we propose several recommendations for
different GEC scenarios. The appropriate repetition penalty value
depends primarily on whether the model has been fine-tuned for the GEC
task: fine-tuned models exhibit their F\textsubscript{0.5} optima in a
narrow range close to 1.0, whereas non--fine-tuned models overcorrect
more strongly and benefit from lower penalty values.

For systems that favor higher precision, such as language-learning
assistance tools or fully automatic systems without human supervision,
values below \textbf{1} should be used. For fine-tuned models, the
analysis in Section~\ref{sec:precision-analysis} shows that values in
the range \textbf{0.95--0.99} coincided with the
F\textsubscript{0.5} optimum across all three main benchmarks, providing
precision improvements at negligible F\textsubscript{0.5} cost. For
non--fine-tuned models, stronger penalties of \textbf{0.90--0.97} were
optimal -- since in most cases the optimum lay at the boundary of the
tested range, even lower values may be beneficial, although they
require per-case evaluation, as both precision and recall may
eventually decrease.

For systems aimed at higher recall, such as book or formal text
proofreading by natives where the primary goal is to correct as many
errors as possible, values above \textbf{1} increase recall at the cost
of precision. Unlike the precision-oriented setting, no single value
can be recommended: the standard benchmarks define no recall-weighted
target metric, and the appropriate weighting of recall against
precision is application-dependent. Nevertheless, the range
\textbf{1.01--1.05} can be perceived as a safe starting point for
increasing recall, as within it recall rises consistently while both
metrics rarely degrade simultaneously; beyond \textbf{1.05}, precision
and recall may deteriorate jointly, and such values require additional
evaluation.

We did not conduct detailed experiments with values lower than
\textbf{0.90} or higher than \textbf{1.10}; however, our initial
findings suggest that such extreme values are more likely to degrade
system performance and introduce strong biases resulting in unnatural
model behavior.

\section{Conclusion}

In this work, we analyze inference-time decoding strategies for large language models in the context of grammatical error correction. Our results show that the selection and configuration of inference parameters are task-dependent and can be effectively adapted to meet specific task requirements without modifying model training or prompting.

Building on the prior work on precision–recall trade-off adaptation for GEC systems with different architectures, such as Align-and-Predict Decoding, we demonstrate that the precision–recall trade-off can likewise be controlled at inference time for LLM-based GEC, using only standard decoding parameters. In particular, our experiments show that the repetition penalty parameter provides a simple and effective mechanism for adjusting the precision–recall balance during decoding.

This work opens several directions for future research, including extending the analysis to additional LLMs, languages, and domains, as well as exploring the applicability of inference-time control mechanisms to related text editing tasks such as text simplification.

\section*{Limitations}

This work explores inference-time decoding strategies for LLM-based GEC; however, the availability of decoding parameters varies between APIs and deployment frameworks. We do not evaluate proprietary models such as GPT-4 or GPT-5, as they currently do not expose a repetition penalty parameter.

The repetition penalty parameter behaves differently across models and benchmarks; therefore, it must be carefully selected for each specific situation and context.

Our experiments focus on three LLMs evaluated under three configurations, which may limit the generalizability of our findings across models of different sizes or architectures.

Finally, we evaluate our approach only on English, German, and Ukrainian GEC benchmarks. Extending the analysis to a larger number of languages would help better assess whether inference-time penalty behavior differs across languages.

\bibliography{custom}

\appendix

\section{Training \& Inference Details}
\label{sec:appendix}

We trained our models on a 1–4×A100 (80GB) GPU setup, depending on the model size, and evaluated them on either a 2×RTX 3090 GPU setup or a 1×A100 (80GB) setup. The models were trained on the training portions of the datasets. Training a single model took around one hour. The models were trained separately for each language used in the experiments. The total inference time was approximately 2–3 hours. For inference, we used vLLM in version 0.9.1, however the repetition mechanism is still valid and unchanged in newer vLLM versions. The hyperparameter values are described in Table~\ref{tab:hyperparameters}. The following prompt was used during training our models and during inference:

\begin{lstlisting}[language={}, numbers=none, frame=single, breaklines=true, basicstyle=\ttfamily\scriptsize]
You are a professional grammar and language correction system.
Your task is to correct all errors in the given text.
Do not change the original meaning, tone, or style.
Do not add new information or remove existing content unless required for grammatical correctness.

Output requirements:

Return only the corrected version of the text.

Do not include explanations, comments, annotations, or formatting changes.

Do not restate the original text.

Text to correct:

<text to correct>
\end{lstlisting}

Listing~\ref{lst:listing} shows two options for controlling the repetition penalty value during LLM inference: directly through vLLM's \texttt{SamplingParams} and through an OpenAI-compatible API endpoint. In both cases, applying the penalty requires only setting a single parameter, without any modification of the decoding algorithm.

\begin{table}
  \centering
  \begin{tabular}{c|c}
    \hline
    \textbf{Hyperparameter name} & \textbf{Value}  \\ \hline
   learning rate & 5e-6  \\
   total batch size & 16  \\
   warmup steps & 50 \\
   lr scheduler & linear \\
   epochs & 1 \\
   optimizer & AdamW8bit \\
   weight decay & 0.01 \\
   \hline
  \end{tabular}
  \caption{Hyperparameter values used to train our models.}
  \label{tab:hyperparameters}
\end{table}

\begin{lstlisting}[float=tb, numbers=none, frame=none, caption={Python code that allows control over repetition penalty values.}, captionpos=b, label={lst:listing}]
# Option 1
sampling_params = SamplingParams(n=1, temperature=0, max_tokens=2000, repetition_penalty=1.05)
output = model.generate(input_texts, sampling_params)

# Option 2
client.chat.completions.create(
    model=model_name,
    messages=messages,
    temperature=0.0,
    max_tokens=2000,
    extra_body={
        "repetition_penalty": 1.05
    }
)
\end{lstlisting}

\begin{table*}
  \centering
  \begin{tabular}{l|c|c|c|c}
    \textbf{Benchmark} & \textbf{Lang} & \textbf{Split} & \textbf{\# examples} & \textbf{\# references} \\
    \hline
    \textbf{JFLEG}          & en & test  & 747  & 4\\ \hline
    \textbf{CoNLL-2014}     & en & test  & 1.3k & 2 \\ \hline
    \textbf{W\&I-LOCNESS}   & en & train   & 34.3k & 1 \\
    \textbf{W\&I-LOCNESS}   & en & test  & 4.5k & 5 \\ \hline
    \textbf{UA-GEC}   & uk & train   & 31.0k & 1 \\
    \textbf{UA-GEC}   & uk & test  & 2.7k & 2 \\ \hline
    \textbf{Falko-MERLIN}   & de & train   & 19.2k & 1 \\
    \textbf{Falko-MERLIN}   & de & test  & 2.3k & 1 \\ \hline
  \end{tabular}
  \caption{Statistics on GEC datasets used in our work. \# References indicates the number of reference corrections provided per input sentence.}
  \label{tab:dataset_stats}
\end{table*}

\begin{table*}[t]
\centering
\small
\begin{tabular}{lcccccccc}
\toprule
\multirow{2}{*}{\textbf{Model}} & \multirow{2}{*}{\textbf{Rep. Penalty}} &
\multicolumn{3}{c}{\textbf{CoNLL-2014-test}} &
\multicolumn{3}{c}{\textbf{BEA-test}} &
\textbf{JFLEG} \\
\cmidrule(lr){3-5} \cmidrule(lr){6-8} \cmidrule(lr){9-9}
& & P & R & F\textsubscript{0.5} & P & R & F\textsubscript{0.5} & GLEU \\
\midrule
Qwen3-4B-Prompted & 0.90 & \textbf{57.32} & 52.05 & \textbf{56.18} & \textbf{53.82} & 62.69 & \textbf{55.39} & 61.34 \\
 & 0.95 & 56.13 & 53.64 & 55.61 & 52.47 & 63.74 & 54.40 & 61.81 \\
 & 0.97 & 55.46 & 53.78 & 55.12 & 51.67 & 64.18 & 53.77 & 62.04 \\
 & 0.99 & 54.77 & 53.96 & 54.60 & 51.14 & 64.24 & 53.31 & 61.94 \\
 & 1.00 & 54.75 & 54.14 & 54.63 & 50.85 & 64.41 & 53.09 & 61.95 \\
 & 1.01 & 54.21 & 54.46 & 54.26 & 50.52 & 64.62 & 52.83 & 61.83 \\
 & 1.03 & 53.85 & 54.75 & 54.03 & 49.76 & 64.89 & 52.19 & \textbf{62.16} \\
 & 1.05 & 53.20 & 55.12 & 53.58 & 48.82 & 64.96 & 51.37 & 62.13 \\
 & 1.10 & 51.28 & \textbf{55.72} & 52.11 & 46.46 & \textbf{65.23} & 49.29 & 61.90 \\
\midrule
Qwen3-4B-FT-Prompted & 0.90 & 69.25 & 38.65 & 59.79 & 71.14 & 51.37 & 66.06 & 58.58 \\
 & 0.95 & \textbf{70.62} & 45.74 & 63.69 & \textbf{72.50} & 59.61 & 69.49 & 61.03 \\
 & 0.97 & 70.20 & 47.89 & \textbf{64.22} & 71.80 & 62.22 & \textbf{69.65} & 61.83 \\
 & 0.99 & 68.96 & 49.45 & 63.92 & 70.48 & 64.27 & 69.15 & 62.40 \\
 & 1.00 & 68.06 & 50.21 & 63.54 & 69.57 & 65.00 & 68.61 & 62.53 \\
 & 1.01 & 67.29 & 50.88 & 63.21 & 68.54 & 66.05 & 68.03 & \textbf{62.54} \\
 & 1.03 & 65.05 & 53.35 & 62.32 & 65.95 & 67.32 & 66.22 & 62.34 \\
 & 1.05 & 62.68 & 55.19 & 61.02 & 62.22 & 68.08 & 63.31 & 61.94 \\
 & 1.10 & 54.24 & \textbf{58.60} & 55.06 & 48.66 & \textbf{69.03} & 51.71 & 59.12 \\
\midrule
Qwen3-4B-FT-NoPrompt & 0.90 & 67.69 & 41.19 & 59.97 & 72.17 & 55.91 & 68.20 & 59.27 \\
 & 0.95 & \textbf{69.55} & 47.11 & 63.50 & \textbf{72.25} & 61.33 & 69.77 & 61.60 \\
 & 0.97 & 68.56 & 48.10 & 63.18 & 71.93 & 63.15 & \textbf{69.98} & 62.28 \\
 & 0.99 & 68.10 & 50.00 & 63.50 & 70.84 & 64.59 & 69.50 & 62.91 \\
 & 1.00 & 67.75 & 50.94 & \textbf{63.55} & 70.11 & 65.46 & 69.13 & 63.02 \\
 & 1.01 & 66.93 & 51.29 & 63.08 & 69.08 & 65.95 & 68.43 & \textbf{63.16} \\
 & 1.03 & 65.05 & 52.71 & 62.14 & 67.32 & 66.66 & 67.19 & 63.14 \\
 & 1.05 & 62.08 & 54.04 & 60.28 & 64.19 & 67.55 & 64.83 & 62.84 \\
 & 1.10 & 54.63 & \textbf{57.60} & 55.20 & 51.57 & \textbf{67.87} & 54.17 & 60.34 \\
\bottomrule
\end{tabular}
\caption{Effect of the repetition penalty on English GEC performance for Qwen3-4B models. A repetition penalty value of 1 indicates that no penalty is applied.}
\label{tab:rep_penalty_conll_bea_jfleg_qwen}
\end{table*}

\begin{table*}[t]
\centering
\small
\begin{tabular}{lcccccccc}
\toprule
\multirow{2}{*}{\textbf{Model}} & 
\textbf{Repetition} &
\multicolumn{3}{c}{\textbf{CoNLL-2014-test}} &
\multicolumn{3}{c}{\textbf{BEA-test}} &
\textbf{JFLEG} \\
\cmidrule(lr){3-5} \cmidrule(lr){6-8} \cmidrule(lr){9-9}
& \textbf{Penalty} &
P & R & F\textsubscript{0.5} &
P & R & F\textsubscript{0.5} &
GLEU \\
\midrule
Gemma2-9B-Prompted & 0.90 & \textbf{54.70} & 56.66 & \textbf{55.08} & \textbf{48.44} & 68.38 & \textbf{51.44} & 64.03 \\
 & 0.95 & 53.71 & 57.20 & 54.37 & 47.14 & 68.99 & 50.33 & \textbf{64.06} \\
 & 0.97 & 53.33 & 57.75 & 54.16 & 46.52 & 69.28 & 49.79 & 63.95 \\
 & 0.99 & 52.94 & 58.24 & 53.92 & 45.89 & \textbf{69.47} & 49.23 & 63.89 \\
 & 1.00 & 52.75 & 58.40 & 53.79 & 45.39 & 69.42 & 48.77 & 63.80 \\
 & 1.01 & 52.86 & 58.83 & 53.95 & 44.92 & 69.24 & 48.31 & 63.73 \\
 & 1.03 & 52.37 & 58.89 & 53.56 & 43.93 & 69.02 & 47.37 & 63.45 \\
 & 1.05 & 51.72 & 58.91 & 53.02 & 42.86 & 68.84 & 46.36 & 63.33 \\
 & 1.10 & 50.19 & \textbf{59.73} & 51.85 & 40.36 & 68.31 & 43.96 & 62.63 \\
\midrule
Gemma2-9B-FT-Prompted & 0.90 & 72.44 & 46.61 & 65.21 & 75.75 & 62.09 & 72.56 & 60.78 \\
 & 0.95 & 73.25 & 52.66 & 67.94 & \textbf{76.41} & 68.02 & 74.57 & 62.79 \\
 & 0.97 & \textbf{73.48} & 54.51 & 68.70 & 76.12 & 69.98 & \textbf{74.81} & 63.46 \\
 & 0.99 & 72.67 & 56.68 & 68.79 & 75.01 & 71.88 & 74.36 & 64.00 \\
 & 1.00 & 72.57 & 57.53 & \textbf{68.96} & 74.76 & 72.50 & 74.30 & 64.19 \\
 & 1.01 & 72.15 & 58.50 & 68.93 & 73.74 & 73.12 & 73.62 & \textbf{64.23} \\
 & 1.03 & 69.43 & 60.46 & 67.43 & 71.16 & 73.97 & 71.70 & 63.98 \\
 & 1.05 & 67.14 & 61.59 & 65.95 & 68.13 & \textbf{74.23} & 69.27 & 63.55 \\
 & 1.10 & 59.13 & \textbf{63.24} & 59.91 & 57.20 & 74.03 & 59.93 & 61.20 \\
 \midrule
Gemma2-9B-FT-NoPrompt & 0.90 & 72.42 & 43.05 & 63.73 & 76.39 & 57.90 & 71.80 & 59.94 \\
 & 0.95 & 73.75 & 51.28 & 67.81 & \textbf{77.33} & 66.63 & 74.93 & 62.59 \\
 & 0.97 & \textbf{73.96} & 53.82 & 68.81 & 76.82 & 69.62 & \textbf{75.26} & 63.40 \\
 & 0.99 & 73.35 & 56.81 & \textbf{69.31} & 75.44 & 72.02 & 74.73 & 64.24 \\
 & 1.00 & 72.60 & 57.66 & 69.03 & 74.55 & 72.80 & 74.19 & \textbf{64.32} \\
 & 1.01 & 71.64 & 58.57 & 68.58 & 73.80 & 73.27 & 73.69 & \textbf{64.32} \\
 & 1.03 & 69.29 & 60.84 & 67.41 & 70.58 & 74.53 & 71.33 & 64.07 \\
 & 1.05 & 66.29 & 62.62 & 65.52 & 67.21 & \textbf{74.87} & 68.62 & 63.57 \\
 & 1.10 & 57.76 & \textbf{64.32} & 58.96 & 54.85 & 74.48 & 57.91 & 60.08 \\
\bottomrule
\end{tabular}
\caption{Effect of the repetition penalty on English GEC performance for Gemma2-9B models. A repetition penalty value of 1 indicates that no penalty is applied.}
\label{tab:rep_penalty_conll_bea_jfleg_gemma2}
\end{table*}

\begin{table*}[t]
\centering
\small
\begin{tabular}{lcccccccc}
\toprule
\multirow{2}{*}{\textbf{Model}} & \multirow{2}{*}{\textbf{Rep. Penalty}} &
\multicolumn{3}{c}{\textbf{CoNLL-2014-test}} &
\multicolumn{3}{c}{\textbf{BEA-test}} &
\textbf{JFLEG} \\
\cmidrule(lr){3-5} \cmidrule(lr){6-8} \cmidrule(lr){9-9}
& & P & R & F\textsubscript{0.5} & P & R & F\textsubscript{0.5} & GLEU \\
\midrule
EuroLLM-22B-Prompted & 0.90 & 59.73 & 47.27 & 56.74 & 57.41 & 59.00 & 57.72 & 60.29 \\
 & 0.95 & \textbf{60.79} & 51.76 & 58.74 & \textbf{57.67} & 64.33 & \textbf{58.89} & 62.92 \\
 & 0.97 & 60.19 & 53.89 & \textbf{58.81} & 55.98 & 65.95 & 57.72 & 63.68 \\
 & 0.99 & 57.89 & 55.90 & 57.48 & 53.12 & 66.99 & 55.41 & \textbf{63.92} \\
 & 1.00 & 56.87 & 56.59 & 56.82 & 51.52 & 67.09 & 54.03 & 63.89 \\
 & 1.01 & 55.17 & 57.42 & 55.61 & 49.39 & 67.18 & 52.16 & 63.44 \\
 & 1.03 & 51.98 & 58.50 & 53.16 & 44.57 & \textbf{67.66} & 47.83 & 63.12 \\
 & 1.05 & 48.54 & 59.29 & 50.37 & 38.47 & 66.86 & 42.04 & 61.34 \\
 & 1.10 & 39.07 & \textbf{60.83} & 42.08 & 23.32 & 63.38 & 26.70 & 50.49 \\
\midrule
EuroLLM-22B-FT-Prompted & 0.90 & 66.08 & 32.01 & 54.48 & 71.34 & 42.00 & 62.59 & 53.30 \\
 & 0.95 & 71.72 & 44.35 & 63.84 & 74.23 & 57.75 & 70.23 & 60.67 \\
 & 0.97 & \textbf{72.40} & 48.93 & 66.06 & \textbf{74.37} & 63.72 & 71.96 & 62.39 \\
 & 0.99 & 71.44 & 53.64 & \textbf{66.99} & 74.28 & 68.13 & \textbf{72.96} & 63.82 \\
 & 1.00 & 70.49 & 55.31 & 66.82 & 72.93 & 69.70 & 72.26 & 63.88 \\
 & 1.01 & 69.52 & 57.50 & 66.73 & 70.49 & 71.04 & 70.60 & \textbf{64.03} \\
 & 1.03 & 64.23 & 59.84 & 63.30 & 64.95 & 72.38 & 66.31 & 63.55 \\
 & 1.05 & 58.71 & 62.26 & 59.39 & 57.31 & \textbf{72.74} & 59.85 & 61.20 \\
 & 1.10 & 42.82 & \textbf{64.67} & 45.92 & 30.26 & 67.31 & 34.00 & 48.07 \\
\midrule
EuroLLM-22B-FT-NoPrompt & 0.90 & 67.42 & 37.99 & 58.37 & 72.06 & 51.80 & 66.83 & 55.42 \\
 & 0.95 & 70.84 & 46.63 & 64.18 & 74.41 & 62.19 & 71.59 & 61.52 \\
 & 0.97 & \textbf{71.62} & 49.82 & 65.86 & \textbf{74.68} & 65.69 & 72.69 & 62.54 \\
 & 0.99 & 71.15 & 52.80 & \textbf{66.53} & 74.47 & 68.53 & \textbf{73.20} & 63.49 \\
 & 1.00 & 70.31 & 54.46 & 66.44 & 73.24 & 70.00 & 72.57 & 63.67 \\
 & 1.01 & 69.30 & 56.27 & 66.23 & 71.48 & 71.03 & 71.39 & \textbf{64.01} \\
 & 1.03 & 65.27 & 58.50 & 63.79 & 66.47 & 71.62 & 67.44 & 63.68 \\
 & 1.05 & 60.67 & 60.67 & 60.67 & 59.73 & \textbf{72.15} & 61.86 & 62.26 \\
 & 1.10 & 44.68 & \textbf{63.22} & 47.46 & 34.10 & 67.21 & 37.83 & 51.72 \\
\bottomrule
\end{tabular}
\caption{Effect of the repetition penalty on English GEC performance for EuroLLM-22B models. A repetition penalty value of 1 indicates that no penalty is applied.}
\label{tab:rep_penalty_conll_bea_jfleg_eurollm}
\end{table*}

\begin{table*}[t]
\centering
\small
\begin{tabular}{lccccccc}
\toprule
\multirow{2}{*}{\textbf{Model}} & \textbf{Repetition} & \multicolumn{3}{c}{\textbf{Falko-MERLIN}} & \multicolumn{3}{c}{\textbf{UA-GEC}} \\
\cmidrule(lr){3-5} \cmidrule(lr){6-8}
& \textbf{Penalty} & P & R & F\textsubscript{0.5} & P & R & F\textsubscript{0.5} \\
\midrule
Qwen3-4B-Prompted & 0.90 & \textbf{50.83} & 42.02 & \textbf{48.78} & \textbf{35.81} & 38.71 & \textbf{36.36} \\
 & 0.95 & 49.72 & 44.70 & 48.63 & 32.84 & 41.38 & 34.26 \\
 & 0.97 & 49.05 & 45.59 & 48.32 & 31.69 & 42.63 & 33.40 \\
 & 0.99 & 48.06 & 46.17 & 47.67 & 30.90 & 43.69 & 32.82 \\
 & 1.00 & 47.84 & 46.46 & 47.56 & 30.32 & 44.36 & 32.37 \\
 & 1.01 & 47.38 & 46.83 & 47.27 & 29.70 & 45.11 & 31.88 \\
 & 1.03 & 46.10 & 46.99 & 46.28 & 28.06 & 46.06 & 30.44 \\
 & 1.05 & 44.89 & 47.13 & 45.32 & 26.15 & 46.68 & 28.67 \\
 & 1.10 & 41.03 & \textbf{47.76} & 42.22 & 21.17 & \textbf{47.76} & 23.83 \\
\midrule
Qwen3-4B-FT-Prompted & 0.90 & 59.70 & 36.37 & 52.91 & \textbf{82.04} & 32.48 & 62.86 \\
 & 0.95 & \textbf{60.92} & 43.51 & 56.40 & 79.86 & 37.29 & \textbf{65.01} \\
 & 0.97 & 60.91 & 46.32 & 57.30 & 77.40 & 38.74 & 64.52 \\
 & 0.99 & 60.43 & 48.48 & \textbf{57.59} & 75.42 & 41.30 & 64.73 \\
 & 1.00 & 59.96 & 49.46 & 57.52 & 74.16 & 41.90 & 64.27 \\
 & 1.01 & 59.54 & 50.18 & 57.40 & 72.52 & 43.07 & 63.80 \\
 & 1.03 & 58.23 & 51.77 & 56.81 & 68.59 & 45.47 & 62.26 \\
 & 1.05 & 56.23 & 52.85 & 55.52 & 64.23 & 47.42 & 59.98 \\
 & 1.10 & 45.94 & \textbf{53.24} & 47.24 & 50.55 & \textbf{52.74} & 50.97 \\
\midrule
Qwen3-4B-FT-NoPrompt & 0.90 & 60.04 & 35.53 & 52.76 & \textbf{79.84} & 28.84 & 58.98 \\
 & 0.95 & \textbf{60.92} & 42.77 & 56.15 & 78.84 & 35.11 & 63.12 \\
 & 0.97 & 60.56 & 45.20 & 56.71 & 77.12 & 37.72 & 63.79 \\
 & 0.99 & 60.12 & 47.43 & \textbf{57.06} & 75.62 & 40.92 & \textbf{64.65} \\
 & 1.00 & 59.72 & 48.18 & 56.99 & 74.18 & 42.41 & 64.51 \\
 & 1.01 & 59.30 & 49.32 & 56.99 & 73.24 & 43.74 & 64.54 \\
 & 1.03 & 57.70 & 51.00 & 56.22 & 68.41 & 46.01 & 62.34 \\
 & 1.05 & 55.54 & 51.82 & 54.76 & 64.09 & 48.62 & 60.26 \\
 & 1.10 & 46.39 & \textbf{52.21} & 47.45 & 48.00 & \textbf{53.83} & 49.06 \\
\bottomrule
\end{tabular}
\caption{Effect of the repetition penalty on German/Ukrainian GEC performance for Qwen3-4B models. A repetition penalty value of 1 indicates that no penalty is applied.}
\label{tab:rep_penalty_de_ua_qwen3}
\end{table*}

\begin{table*}[t]
\centering
\small
\begin{tabular}{lccccccc}
\toprule
\multirow{2}{*}{\textbf{Model}} & \textbf{Repetition} & \multicolumn{3}{c}{\textbf{Falko-MERLIN}} & \multicolumn{3}{c}{\textbf{UA-GEC}} \\
\cmidrule(lr){3-5} \cmidrule(lr){6-8}
& \textbf{Penalty} & P & R & F\textsubscript{0.5} & P & R & F\textsubscript{0.5} \\
\midrule
Gemma2-9B-Prompted & 0.90 & \textbf{51.14} & 51.37 & \textbf{51.19} & \textbf{30.67} & 39.89 & \textbf{32.16} \\
 & 0.95 & 49.95 & 53.08 & 50.55 & 29.04 & 43.19 & 31.08 \\
 & 0.97 & 49.43 & 53.68 & 50.22 & 27.88 & 43.99 & 30.08 \\
 & 0.99 & 48.63 & 54.43 & 49.69 & 26.92 & 45.28 & 29.30 \\
 & 1.00 & 48.31 & 54.87 & 49.49 & 26.29 & 45.67 & 28.73 \\
 & 1.01 & 47.47 & 54.64 & 48.75 & 25.87 & 46.97 & 28.43 \\
 & 1.03 & 46.55 & \textbf{55.04} & 48.03 & 24.93 & 49.00 & 27.65 \\
 & 1.05 & 45.30 & 54.90 & 46.94 & 23.81 & 50.02 & 26.59 \\
 & 1.10 & 42.43 & 54.83 & 44.44 & 20.63 & \textbf{51.74} & 23.45 \\
\midrule
Gemma2-9B-FT-Prompted & 0.90 & 67.84 & 46.88 & 62.27 & \textbf{86.16} & 43.26 & 71.90 \\
 & 0.95 & 69.54 & 56.42 & 66.45 & 85.82 & 50.54 & 75.31 \\
 & 0.97 & \textbf{69.77} & 59.36 & \textbf{67.40} & 85.40 & 53.41 & \textbf{76.26} \\
 & 0.99 & 69.00 & 61.50 & 67.35 & 83.22 & 56.77 & 76.13 \\
 & 1.00 & 68.74 & 62.34 & 67.35 & 81.98 & 57.86 & 75.67 \\
 & 1.01 & 67.89 & 63.06 & 66.87 & 80.51 & 59.77 & 75.29 \\
 & 1.03 & 66.36 & 64.75 & 66.03 & 76.25 & 62.19 & 72.96 \\
 & 1.05 & 63.90 & \textbf{64.86} & 64.09 & 71.49 & 64.92 & 70.07 \\
 & 1.10 & 55.88 & 64.16 & 57.36 & 56.47 & \textbf{69.17} & 58.62 \\
\midrule
Gemma2-9B-FT-NoPrompt & 0.90 & 68.74 & 46.80 & 62.85 & 84.96 & 39.19 & 68.87 \\
 & 0.95 & 69.66 & 56.37 & 66.52 & \textbf{86.19} & 49.23 & 74.94 \\
 & 0.97 & \textbf{69.98} & 59.61 & 67.63 & 84.32 & 52.70 & 75.28 \\
 & 0.99 & 69.45 & 61.92 & \textbf{67.80} & 82.75 & 56.83 & \textbf{75.83} \\
 & 1.00 & 68.33 & 62.60 & 67.10 & 81.35 & 58.20 & 75.36 \\
 & 1.01 & 67.96 & 63.25 & 66.96 & 80.02 & 59.87 & 74.98 \\
 & 1.03 & 66.46 & 64.82 & 66.13 & 75.57 & 63.10 & 72.70 \\
 & 1.05 & 64.00 & \textbf{65.07} & 64.21 & 70.97 & 65.55 & 69.82 \\
 & 1.10 & 55.10 & 64.05 & 56.68 & 55.95 & \textbf{69.33} & 58.20 \\
\bottomrule
\end{tabular}
\caption{Effect of the repetition penalty on German/Ukrainian GEC performance for Gemma2-9B models. A repetition penalty value of 1 indicates that no penalty is applied.}
\label{tab:rep_penalty_de_ua_gemma2}
\end{table*}

\begin{table*}[t]
\centering
\small
\begin{tabular}{lccccccc}
\toprule
\multirow{2}{*}{\textbf{Model}} & \textbf{Repetition} & \multicolumn{3}{c}{\textbf{Falko-MERLIN}} & \multicolumn{3}{c}{\textbf{UA-GEC}} \\
\cmidrule(lr){3-5} \cmidrule(lr){6-8}
& \textbf{Penalty} & P & R & F\textsubscript{0.5} & P & R & F\textsubscript{0.5} \\
\midrule
EuroLLM-22B-Prompted & 0.90 & \textbf{53.13} & 45.83 & 51.49 & \textbf{45.01} & 35.71 & \textbf{42.78} \\
 & 0.95 & 52.71 & 52.31 & 52.63 & 40.25 & 41.32 & 40.46 \\
 & 0.97 & 52.34 & 54.55 & \textbf{52.77} & 38.11 & 44.55 & 39.25 \\
 & 0.99 & 51.08 & 55.97 & 51.99 & 34.59 & 47.57 & 36.59 \\
 & 1.00 & 50.19 & 56.34 & 51.31 & 33.26 & 49.37 & 35.58 \\
 & 1.01 & 48.97 & 56.65 & 50.34 & 31.62 & 51.05 & 34.23 \\
 & 1.03 & 46.39 & \textbf{57.09} & 48.19 & 27.76 & 52.82 & 30.67 \\
 & 1.05 & 42.71 & 56.56 & 44.91 & 24.71 & 55.92 & 27.81 \\
 & 1.10 & 30.89 & 52.38 & 33.65 & 16.56 & \textbf{57.85} & 19.31 \\
\midrule
EuroLLM-22B-FT-Prompted & 0.90 & 64.98 & 41.34 & 58.31 & \textbf{83.61} & 35.03 & 65.45 \\
 & 0.95 & 68.07 & 53.10 & 64.44 & 82.96 & 44.53 & 70.75 \\
 & 0.97 & \textbf{68.27} & 55.97 & 65.40 & 81.16 & 48.01 & 71.31 \\
 & 0.99 & 67.73 & 58.77 & \textbf{65.72} & 79.81 & 52.08 & \textbf{72.12} \\
 & 1.00 & 67.28 & 59.92 & 65.67 & 78.19 & 53.42 & 71.56 \\
 & 1.01 & 66.44 & 60.80 & 65.23 & 75.33 & 55.23 & 70.22 \\
 & 1.03 & 64.35 & 62.71 & 64.01 & 71.00 & 58.14 & 67.99 \\
 & 1.05 & 60.87 & \textbf{63.07} & 61.30 & 66.09 & 60.99 & 65.00 \\
 & 1.10 & 49.33 & 62.39 & 51.48 & 48.72 & \textbf{66.75} & 51.50 \\
\midrule
EuroLLM-22B-FT-NoPrompt & 0.90 & 64.82 & 40.22 & 57.76 & \textbf{84.74} & 31.30 & 63.17 \\
 & 0.95 & 67.45 & 51.44 & 63.50 & 83.29 & 41.91 & 69.56 \\
 & 0.97 & \textbf{67.50} & 55.09 & 64.59 & 82.42 & 47.77 & \textbf{71.98} \\
 & 0.99 & 66.81 & 57.77 & \textbf{64.78} & 79.45 & 51.31 & 71.60 \\
 & 1.00 & 65.87 & 59.14 & 64.40 & 78.40 & 53.31 & 71.65 \\
 & 1.01 & 64.71 & 59.63 & 63.63 & 75.82 & 55.43 & 70.62 \\
 & 1.03 & 62.42 & 61.43 & 62.22 & 72.13 & 59.05 & 69.07 \\
 & 1.05 & 59.69 & \textbf{62.55} & 60.24 & 66.36 & 62.55 & 65.56 \\
 & 1.10 & 48.06 & 61.74 & 50.29 & 48.32 & \textbf{68.11} & 51.30 \\
\bottomrule
\end{tabular}
\caption{Effect of the repetition penalty on German/Ukrainian GEC performance for EuroLLM-22B models. A repetition penalty value of 1 indicates that no penalty is applied.}
\label{tab:rep_penalty_de_ua_eurollm}
\end{table*}

\end{document}